\documentclass{article}
\usepackage[preprint]{neurips_2026}
\usepackage[utf8]{inputenc}
\usepackage[T1]{fontenc}
\usepackage[hidelinks]{hyperref}
\usepackage{url,booktabs,tabularx,longtable,amsmath,amssymb,microtype,xcolor,graphicx,xspace}
\usepackage{placeins}
\graphicspath{{figs/}}
\newcommand{\sys}{\textsc{LabFactory}\xspace}

\newcommand{\titleicon}{\raisebox{-0.18em}{\includegraphics[height=1.18em]{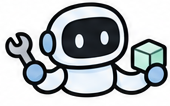}}}
\title{\titleicon\hspace{0.25em}\sys:\\Building and Evaluating Executable AI Labs}
\author{
\normalfont Jinge Wu, Hongjian Zhou, Mingde Zeng, Jiayuan Zhu, Junde Wu,\\
Jiazhen Pan, Lei Clifton\\[4pt]
University of Oxford\quad University College London\\
Technical University of Munich\\
\texttt{jinge.wu@outlook.com}
}

\begin{document}
\maketitle
\begin{abstract}
Scientific tasks specify a desired capability, but realizing it often requires building
a computational system tailored to the task---acquiring data, designing representations,
training models, implementing tools, and deciding how they are used at inference. We
present \sys, a framework in which an AI builder turns a scientific brief into an
executable AI lab: a task-specific solver that integrates models, knowledge resources,
tools, and a controller behind a fixed interface. The builder develops and packages the
lab in a metered workspace; a separate host then executes the delivered artifact on
held-out inputs, with reference labels kept outside the solver's input interface, and
scores its outputs under the task's protocol. This makes the delivered system, rather
than the builder's account of its progress, the object of evaluation. We document 10 selected constructions across
six scientific task categories---from molecular and genomic prediction to
medical imaging, clinical decision support, and biomedical text---whose delivered
labs exceeded their configured reference values on all 12 subtests under host-side
execution. Four contain predictive models fitted during construction; the others assemble
executable analysis environments, knowledge resources, and tool-driven workflows around a
fixed platform LLM. Together they show that an AI agent can carry a scientific brief all
the way to a working lab that can still be invoked, inspected, and checked after
construction ends.

\end{abstract}

\section{Introduction}
\label{sec:v1-intro}
Consider a request to identify functional polyadenylation signals from DNA sequence windows. The requested output is a binary label, but producing it requires a sequence representation, a predictive procedure, and an implementation that can process new inputs. A clinical calculation poses a different construction problem: formulas and unit handling must be implemented exactly, while the parameters they need must be extracted from free-text patient notes. In both settings the scientific brief is only the starting point; the system that turns it into a working computation still has to be built.

Agents that develop models, execute analyses, and run experiments are increasingly studied in machine-learning engineering and scientific workflows \citep{mlagentbench2024,scienceagentbench2025,aiscientistv22025}. We focus on what such activity leaves behind: can an agent assemble the resources and procedures a scientific task requires into a system that remains invocable after construction ends? This shifts attention from individual development actions, or a single answer, to the capabilities retained in the delivered artifact.

\begin{figure}[!ht]
\centering
\includegraphics[width=\linewidth]{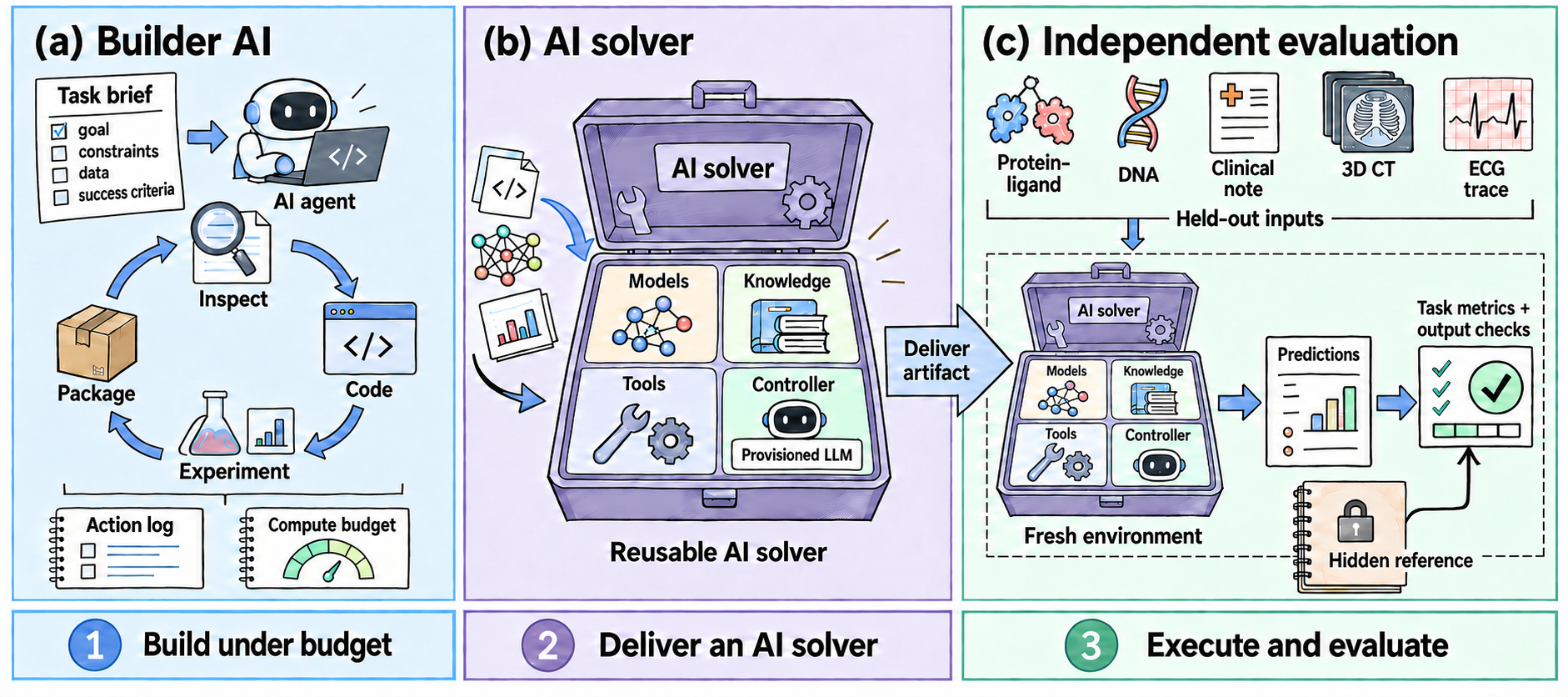}
\caption{\textbf{The \sys framework.} Construction produces a solver composed
of models, knowledge, tools, and a controller. The runtime records actions and
resource consumption; independent evaluation runs the delivered
artifact. ``Hidden reference'' denotes the host-side reference. Illustrated modalities indicate possible task interfaces,
not the reported result set.}
\label{fig:v1-framework}
\end{figure}

We call that artifact an \emph{AI lab}: the task-specific collection of models, knowledge resources, executable tools, and control procedures, delivered as a solver behind a defined interface. \sys separates the builder from the lab it constructs (Figure~\ref{fig:v1-framework}). The builder acquires resources, develops components, and revises its design in a metered workspace. The delivery contract fixes how the lab is invoked but leaves its representations, model families, and inference policy to construction, so different tasks produce different kinds of labs under the same interface. Evaluation is operational: a separate host invokes the delivered lab on held-out inputs and scores its outputs under the task protocol, tying each reported result to an executable system rather than to the builder's own assessment of its progress.

With \texttt{claude-opus-5} as the builder, we report 10 constructions whose delivered labs cleared their configured references under host-side execution, compare three of them with recorded raw-LLM baselines on the same inputs and metrics, and trace the polyadenylation build end to end, from motif-relative alignment and CNN training to an uncertainty-gated inference policy (\S\ref{sec:v1-worked}). Our contributions are:

\begin{enumerate}
\item \textbf{A construction and delivery protocol.} A task-conditioned runtime that records actions, resource use, and terminal status, and a common entry-script contract for independent evaluation (\S\ref{sec:v1-method}).
\item \textbf{An account of constructed AI labs.} A component-level description of delivered solvers, two recurring controller patterns, and construction records for seven labs (\S\ref{sec:v1-composite}, \S\ref{sec:v1-worked}, Appendix~\ref{app:v1-details}).
\item \textbf{Execution-grounded case records.} Host-side results for all 12 subtests with reference provenance, raw-LLM comparisons on the same inputs and metrics, and per-episode construction traces and resource records (\S\ref{sec:v1-cases}, Appendix~\ref{app:expanded-subtests}).
\end{enumerate}

\section{Related Work}
\label{sec:v1-related}
\paragraph{Machine-learning engineering agents.}
MLAgentBench~\citep{mlagentbench2024} and MLE-bench~\citep{mlebench2024} evaluate
agents that prepare data, run experiments, and train models on ML tasks;
MLE-STAR~\citep{mlestar2025} refines model components through web search and
ablation-guided edits. These works establish executable experimentation as an
evaluation unit, typically producing a model or a submission file. In \sys the
product is a solver that may combine trained models with knowledge resources,
tools, and control logic, and that the host executes through a fixed interface.

\paragraph{Agents for scientific workflows.}
ScienceAgentBench~\citep{scienceagentbench2025} evaluates self-contained Python
programs for scientific tasks, and AI Scientist-v2~\citep{aiscientistv22025}
automates hypothesis generation, experimentation, and manuscript writing. The
object here is an intermediate scientific instrument: a solver that remains
invocable on new task inputs after construction has ended.

\paragraph{AI-generated agents.}
The Darwin G\"odel Machine~\citep{dgm2026} modifies agent code and evaluates the
resulting variants empirically. \sys likewise makes an executable system the
product of agent activity, but for a specified scientific task and with an
explicit boundary between the builder and the solver it delivers.

\section{Method}
\label{sec:v1-method}
\label{sec:v1-framework}
\subsection{Task-conditioned construction}
Let a task be a tuple
$(d,\mathcal{X},\mathcal{Y},\mathcal{R},m)$, where $d$ is a scientific
brief, $\mathcal{X}$ and $\mathcal{Y}$ specify input and output formats,
$\mathcal{R}$ defines the available resources and delivery contract, and $m$ is
the evaluation metric. Given $d$ and $\mathcal{R}$, a builder works in a
workspace and delivers an artifact $\mathcal{D}$ (\S\ref{sec:v1-lifecycle}).

The interface specifies what must be delivered without prescribing a single
internal architecture. The builder may deliver a trained predictor, a retrieval
index, an analytical program, or a combination of these, which matters when
moving between tasks whose central operation is prediction, retrieval, or
computation. The framework provides the working and evaluation environments;
representation, model family, tool design, and inference policy are chosen during
construction.

\subsection{The construction lifecycle}
\label{sec:v1-lifecycle}
An episode begins when the builder receives the task brief and a workspace. A
typical run inspects the brief and environment, acquires public data and
software, develops models or tools, validates them on its own development
splits, packages the solver, and submits it. These stages are common rather than
mandatory: after a failed local test the builder can return to an earlier design
decision. Tool responses---a training loss, an exception, a self-check
result---inform the next construction action. They are development feedback,
separate from the independent evaluation that grades the delivered artifact.

An episode ends in one of two ways: the builder submits an artifact, or, if it
stops or times out with a staged artifact but no explicit submission, an operator
closes the episode. This \emph{operator closure} is recorded as a distinct
terminal path, and the staged artifact then proceeds to independent evaluation (\S\ref{sec:v1-eval}).

\subsection{The solver as a composite artifact}
\label{sec:v1-composite}
We describe an artifact as
\begin{equation}
\mathcal{D}=(P,W,K,U,\pi),\qquad \widehat{\mathbf y}=f_{\mathcal{D}}(\mathbf X;F),
\label{eq:v1-artifact}
\end{equation}
where $\mathbf X$ is a batch of task inputs, $\widehat{\mathbf y}$ the corresponding
predictions, $P$ is input processing, $W$ denotes model parameters, $K$ comprises
knowledge and retrieval resources, $U$ is the executable tool interface, and
$\pi$ controls how these components are invoked and combined. $F$ denotes access,
where used, to a platform LLM served through host endpoints. This is a
descriptive decomposition rather than a requirement that every solver contain
every component.

The decomposition separates capabilities that are often conflated in a single
agent prompt: $P$ translates task inputs into stable representations, $W$ can
supply a numerical prediction, $K$ supplies structured domain information, and $U$
exposes operations the controller can execute. The builder does not fine-tune the
platform LLM during construction: it trains domain models and writes the code
through which the LLM is invoked, and the deliverable contains no newly trained
LLM weights.

The controller $\pi$ is the most task-dependent component, and two recurring
controller patterns organize the reported cases (Appendix~\ref{app:v1-details}). A
\emph{fixed-procedure} controller combines a predictor with a prescribed policy
for language-model support: bounded score adjustment for protein interaction,
bounded evidence-guided refinement for protein--ligand affinity, and arbitration within an uncertainty band in the
worked example (\S\ref{sec:v1-worked}). An \emph{LLM-driven loop} has the model
write and run code or queries against constructed tools, as in the
executable-environment cases.

Both share one delivery contract. An artifact is packaged with an entry
script and the code, weights, and resources needed for its configured inference
path. The entry script accepts a batch of cases and produces an
identifier-indexed prediction file, so the host can invoke solvers with different
internals through the same contract while retaining task-specific output types
such as probabilities, continuous estimates, SQL queries, and textual answers.

\subsection{Construction runtime and records}
\label{sec:v1-runtime}
The builder works in a build VM with network access to public datasets, software,
and model weights, and organizes its own development data and local tests. The
runtime admits an action only when the episode is open, the request is well
formed, and the applicable resource ledger allows it; compute jobs reserve budget
and settle against measured use, and model calls pass through host-mediated
endpoints. The ledgers make consumption observable without imposing a single
fixed budget across tasks.

Each run leaves three records. Harness logs retain turns, tool calls, and token
counts; result metadata retain actions, attempts, resource use, and the terminal
path (submission or operator closure); the workspace and deliverable preserve the
code, weights, and resources of the artifact. Together they trace a reported
metric back to a specific construction episode and keep the delivered artifact
separate from the builder's narrative of its own progress.

\subsection{Independent evaluation}
\label{sec:v1-eval}
The host executes the delivered solver on evaluation inputs
$\{x_i\}_{i=1}^n$, obtains outputs $\{\hat y_i\}_{i=1}^n$, and computes the
configured metric $m(\{\hat y_i\},\{y_i\})$ from host-side references, recording
execution status, coverage, and applicable output checks. Evaluation thus ties one
executable artifact to one measured outcome, distinct from the builder's
development measurements. Detailed accounting and grading conventions are given in
Appendix~\ref{app:v1-protocol}.

\paragraph{Data-access boundary.}
\label{sec:v1-boundary}
The host retains evaluation references outside the solver input interface and
records metrics from artifact execution. During construction, control actions pass
through host-mediated endpoints, a separate execution surface runs commands and
retrieves public packages, data, and weights, and structured observations screen
out hidden reference fields. At evaluation, case identifiers are reminted and input
fields permuted, so a solver cannot rely on persistent identifiers or field
positions; these transformations change presentation, not scientific content.
Builders may acquire public resources during construction; the interface boundary
does not establish disjointness between those resources and evaluation samples.

\subsection{Task metrics and acceptance criteria}
\label{sec:v1-metrics}
A task may contain several subtests. For subtest $j$, let $v_j$ denote its
higher-is-better primary value and $b_j$ its configured reference bar.
The acceptance rule is
\begin{equation}
\mathrm{clear}(\mathcal{D})=\bigwedge_{j=1}^{J}(v_j>b_j),\qquad
j^*=\arg\min_j(v_j-b_j).
\label{eq:v1-clear}
\end{equation}
The binding subtest $j^*$ provides a compact task-level row; the raw margin
$v_j-b_j$ is the primary-metric difference, not a reward difference. All subtests
must clear their bars for a case to count as clearing.
The platform also computes a bar-anchored episode reward
for its own bookkeeping; this report uses primary values and reference bars only.
Judge-graded scores are a separate metric type
(Appendix~\ref{app:v1-protocol}) and are not averaged with classification accuracy;
group summaries aggregate only counts and resource use.

\paragraph{Coverage and degenerate outputs.}
The verifier requires coverage of at least half of the current identifiers and
at least $\min(2,n)$ distinct top-1 outputs. Failure zeros the affected subtest;
it need not zero every other subtest. Additional metric-specific rules can
constrain outputs, including positive-rate guards. These checks make incomplete
or constant predictions visible, but are neither semantic-validity certificates
nor proofs of data disjointness.

\section{Results}
\label{sec:v1-cases}
\subsection{Evaluation setup}
\paragraph{Task collection.}
Each construction task is assembled around a published method or benchmark. For
each task we obtain the dataset and hold out a test split on the host. Each task has a configured reference value used for case selection.
Table~\ref{tab:v1-subtests} records the available source attribution, any rescaling
or recomputation, and the provenance status of each value. Before
construction, we checked that each task executes correctly in its evaluation
environment. The builder then constructs an executable solver from the given task
specification within the designated resource environment (\S\ref{sec:v1-runtime});
the held-out split is used only for independent evaluation
(\S\ref{sec:v1-boundary}).

\paragraph{Reported cases.}
A case is reported when its delivered solver cleared the reference bar on every
subtest under host-side execution (Eq.~\eqref{eq:v1-clear}). We report 10
such cases in six categories: molecules and proteins (2), genomics and omics (2),
imaging (1), knowledge and reasoning (2), clinical decision support (1), and
clinical and biomedical text (2) (Table~\ref{tab:v1-groups}). The 10 cases contain
12 subtests with classification, regression, program-synthesis, and judge-scored
outputs; held-out sets range from 285 to 4{,}183 cases per subtest
(Table~\ref{tab:v1-subtests}).

\paragraph{Builder configuration.}
Each case is one construction episode with \texttt{claude-opus-5} as the builder,
run in the pi coding-agent harness; host-mediated endpoints serve the platform
models used at inference. Wall-clock limits were 12 hours in earlier batches and
24 hours in later ones, and resource use was recorded without a common budget
across tasks (Appendix~\ref{app:v1-protocol}).

\paragraph{Scope of the reported results.}
These are demonstrated constructions selected because they cleared their configured
bars; they are not a success rate over the collection. The bars are configured
reference points rather than uniformly matched SOTA comparisons, and their
comparability to each source protocol varies (\S\ref{sec:v1-analysis}). Each case is
a single recorded build under one builder configuration with unequal development
budgets, so the results do not measure build reliability, isolate component
contributions, or support controlled comparisons of performance or resource use
across tasks.

\subsection{Delivered solver artifacts}
\label{sec:v1-artifacts}
Inspection of the 10 delivered artifacts identifies build-fitted predictive models
in four cases: CNN ensembles for polyadenylation signals and 3D organ
classification, a pair predictor on frozen protein-language-model embeddings for
protein interaction, and a boosted-tree ensemble for protein--ligand affinity. The
remaining six primarily assemble tools, knowledge resources, and control procedures
around the platform LLM. These counts refer to reusable parameters fitted during
construction; statistical estimation performed on task inputs is treated as
analysis.

The delivered systems differ in how these parts are combined. Polyadenylation
signal detection combines a CNN sequence predictor with LLM
arbitration in an uncertainty band (\S\ref{sec:v1-worked}); protein--ligand
affinity estimation combines structural descriptors with a fitted tree ensemble
and bounded, evidence-guided refinement. Bioinformatics analysis delivers an
executable scientific workspace with analysis recipes and offline gene-set
resources; clinical text-to-SQL delivers schema search, a schema-faithful
synthetic database, and query-execution checks. Clinical calculation separates
LLM parameter extraction from 55 executable calculator specifications with unit
conversion, and chemistry reasoning supplies molecular and formula tools,
reference knowledge, and answer checks. All share the delivery contract of
\S\ref{sec:v1-framework}; Appendix~\ref{app:v1-details} gives seven representative systems' inputs,
components, and output.

\subsection{Recorded task results}
\begin{figure}[!ht]
\centering
\includegraphics[width=\linewidth]{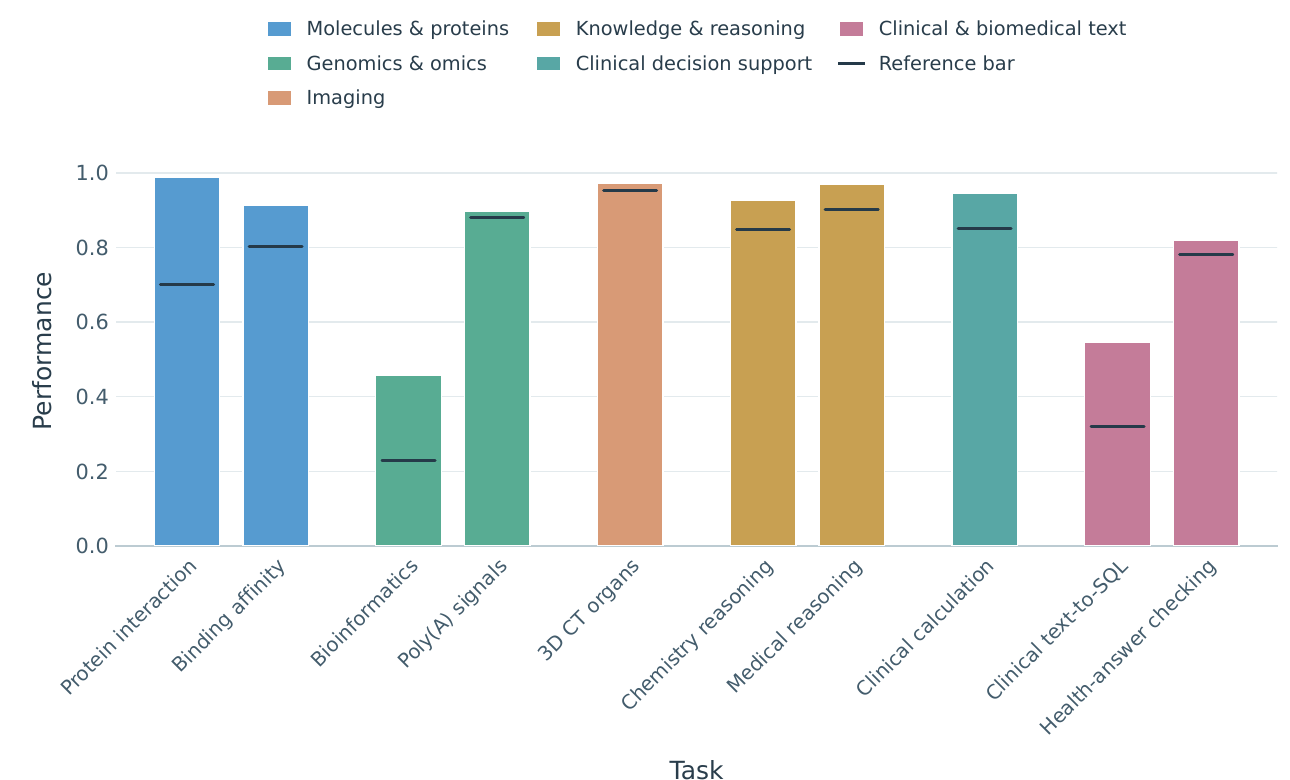}
\caption{\textbf{Performance across 10 scientific tasks.}
Colored columns show host-side results of the delivered artifacts; dark horizontal
ticks mark the task-specific reference bars. Colors distinguish scientific
categories. Scores use heterogeneous metrics on their native scale, so column heights and gaps are not
comparable effect sizes across tasks. Column labels abbreviate the task names of
Table~\ref{tab:v1-subtests}, which lists the tasks in the same order.}
\label{fig:v1-dumbbell}
\end{figure}

Figure~\ref{fig:v1-dumbbell} shows the host-side result of each of the 10 cases
against its reference bar. Each column uses the task's smallest-margin (binding)
subtest. Scores are native task metrics rather than a common effect size, so each
column is compared with its own reference marker. Exact values, sample counts, and bars for all 12 subtests are given in
Appendix~\ref{app:expanded-subtests}, and the comparability of the
reference bars is discussed in \S\ref{sec:v1-analysis}.

\paragraph{Prediction tasks.}
In the molecular, genomic, and imaging categories, the prediction solvers contain a
model fitted during construction and are scored on native prediction metrics.
Protein--protein interaction reaches AUROC 0.9874 on 3{,}000 pairs (reference 0.70),
and protein--ligand affinity Pearson $r$ 0.9141 on the 285 CASF-2016 complexes
(0.803; Table~\ref{tab:v1-subtests}, note~3). Poly(A) signal detection (F1 0.8977
vs.\ 0.88) and 3D abdominal organ classification (accuracy 0.9705 vs.\ 0.952) clear
their references by less than two points. Bioinformatics analysis, which delivers an
executable analysis environment rather than a fitted model, records judge accuracy
0.4561 on 296 questions against a reference of 0.228 (note~1).

\paragraph{Knowledge and reasoning.}
The two knowledge and reasoning cases assemble tools, reference knowledge, and
answer checks around the platform LLM. Medical question answering reaches 0.9678 on
MedQA, 0.8735 on MedMCQA, and 0.9123 on Medbullets, and ChemBench reaches 0.9268.
For these accuracy-scored subtests, the comparison below places the solver results
alongside recorded raw-LLM baselines on the same evaluation inputs.

\paragraph{Clinical decision support and text.}
Clinical calculation records judge accuracy 0.9445 on 1{,}100 notes (strict string
match 0.8755) against a reference of 0.85. Among text tasks, clinical text-to-SQL
reaches execution accuracy 0.5470 on 1{,}000 questions (0.32), and consumer-health
answer verification 0.8193 (0.78).

\paragraph{Comparison with recorded raw-LLM baselines.}
Recorded comparisons include MedMCQA, where the constructed solver achieved
87.35\% accuracy compared with 79.30\% for the recorded raw-LLM baseline, and
MEDIQA, where accuracy increased from 78.68\% to 81.93\%. The complete set of
available accuracy comparisons, covering five subtests from three of the reported cases,
is reported in Appendix~\ref{sec:v1-matched-baselines}. These comparisons match inputs
and metrics, but the archived records do not establish that the baseline and solver
used the same LLM; the differences therefore do not isolate the effect of construction.

\subsection{Resource use}
\label{app:v1-resources}
\begin{figure}[!ht]
\centering
\includegraphics[width=\linewidth]{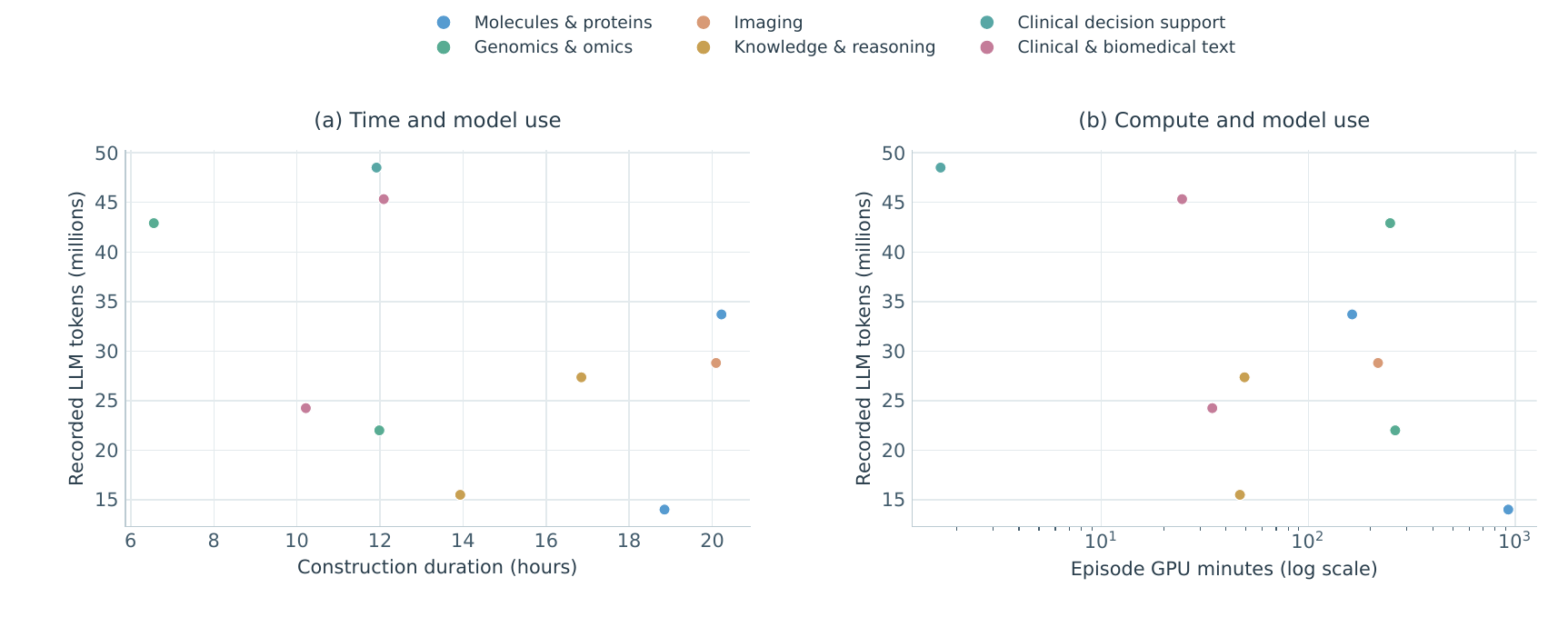}
\caption{\textbf{Recorded time and resource consumption of the 10 selected runs.}
Each point is one episode, and colors identify task groups. These are descriptive
construction-episode measurements, not inference costs.}
\label{fig:v1-resources}
\end{figure}

Hours are elapsed durations from the selected run's harness log; tokens, GPU
minutes, and tool credits are episode-level accounting values for that run, not
inference latency or the total cost over all attempted builds. Tokens are reported
as recorded, without assuming cached, input, and output tokens share a price, and
tool credits are internal accounting units rather than a billed currency. Across
the 10 episodes the medians are 13.01 hours, 28.08 million tokens, 106.1 GPU
minutes, and 2.57 million tool credits, over a median of 216.5 agent turns and 220.5
tool calls per build (144--262 turns across cases). Table~\ref{tab:v1-groups}
gives group medians, and Figure~\ref{fig:v1-resources} shows the episode-level
distributions of duration, model tokens, and GPU use; GPU minutes cover the
complete episode and are not restricted to model training.

\begin{table}[!ht]
\centering\small
\caption{Task-group counts and median selected-run resource use. Each resource
column is a within-group median, not a sum. Imaging and clinical decision support each contain one case. Heterogeneous performance metrics are not averaged.}
\label{tab:v1-groups}
\begin{tabular}{@{}lrrrrr@{}}\toprule
Task group & Cases & Hours & Tokens M & GPU min & Credits M\\\midrule
Molecules \& proteins & 2 & 19.54 & 23.86 & 544.5 & 1.04\\
Genomics \& omics & 2 & 9.27 & 32.45 & 256.0 & 0.75\\
Imaging & 1 & 20.09 & 28.80 & 217.4 & 0.49\\
Knowledge \& reasoning & 2 & 15.39 & 21.43 & 48.0 & 4.13\\
Clinical decision support & 1 & 11.92 & 48.51 & 1.7 & 9.43\\
Clinical \& biomedical text & 2 & 11.15 & 34.79 & 29.5 & 13.38\\
\midrule
All reported cases & 10 & 13.01 & 28.08 & 106.1 & 2.57\\\bottomrule\end{tabular}
\end{table}

\FloatBarrier

\section{Case Study: Polyadenylation Signal Detection}
\label{sec:v1-worked}
The task is to decide whether a candidate polyadenylation signal in a roughly
600-base DNA window is functional; the host evaluates positive-class F1 on 3{,}000
held-out windows against a reference bar of 0.88. We follow the construction
process, the inference path of the delivered solver, and the result of the
complete system.

\subsection{Construction process}
The builder first implements motif-relative canonicalization: each window is
re-centred on its \texttt{AATAAA} occurrence, keeping 294 bases on either side, so
that downstream components see a fixed four-channel, 594-position input. From the
public DeepGSR collection it assembles 22{,}604 human windows and partitions them
before any fitting into 17{,}104 training, 2{,}500 validation, and 3{,}000
development-test windows. On this data it trains dilated residual 1D CNNs with
sequence-shift augmentation, selects four of them greedily on the validation split
into an ensemble, and tunes the decision threshold on the same split for
positive-class F1, arriving at 0.495.

\begin{figure}[!ht]
\centering
\includegraphics[width=\linewidth]{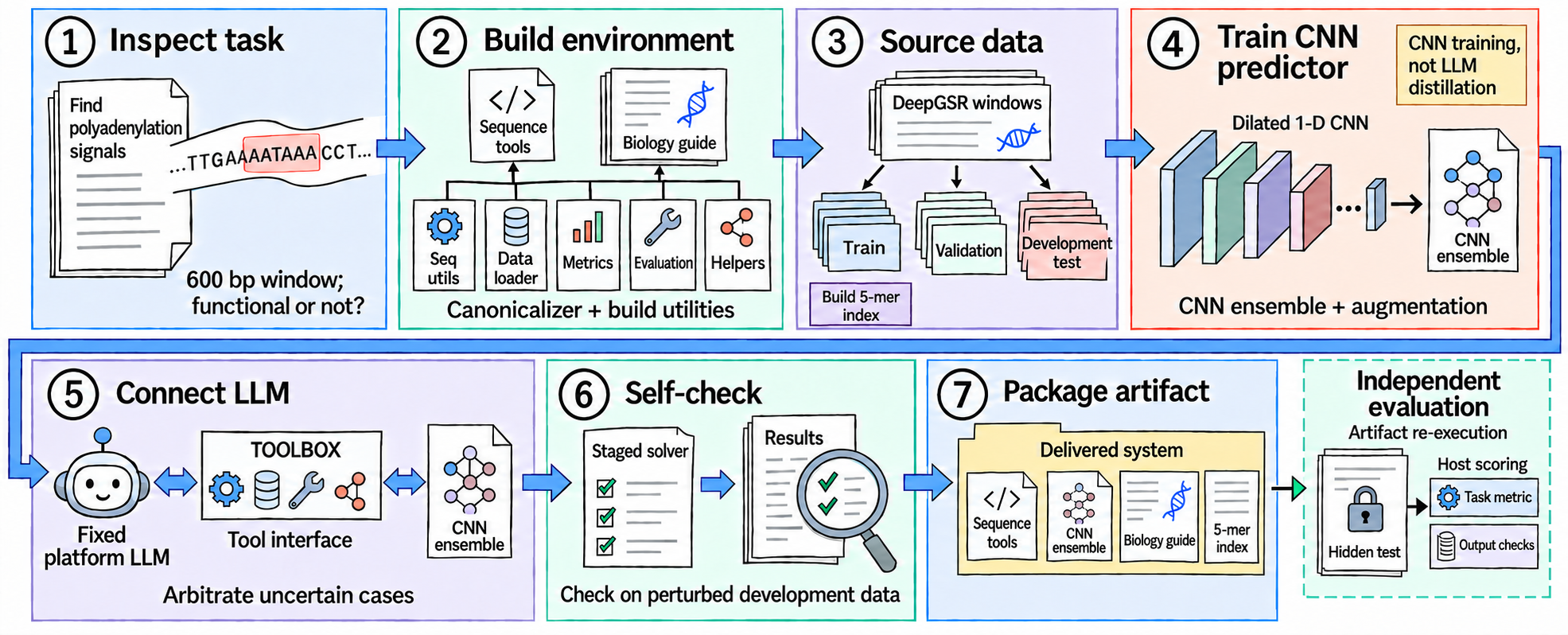}
\caption{\textbf{Construction of the polyadenylation solver.} The builder
implements build utilities and a task interface, obtains data, constructs a
5-mer retrieval index, trains a CNN predictor, connects the LLM,
performs development self-checks, and packages the artifact. The train,
validation, and development-test split belongs to construction; the final box is the evaluation of
\S\ref{sec:v1-eval}.}
\label{fig:v1-builder}
\end{figure}

Around the predictor the builder constructs three further resources: a 5-mer
composition index that returns the nearest labelled reference windows, a biology
knowledge card describing cis-elements around the signal, and a shared tool
interface through which the aligned sequence, sequence features, the CNN score,
retrieved neighbours, and a compact evidence summary are exposed. A driver
connects these tools to the platform LLM through a pi agent and packages the
solver behind the standard entry script. Self-checks run the packaged solver with
fresh identifiers, shifted window boundaries, off-centre signals, and reordered
inputs (Figure~\ref{fig:v1-builder}). The episode ended with a staged artifact
and no explicit submission, so it reached host-side execution through operator
closure.

\subsection{Inference path}
\begin{figure}[!ht]
\centering
\includegraphics[width=\linewidth]{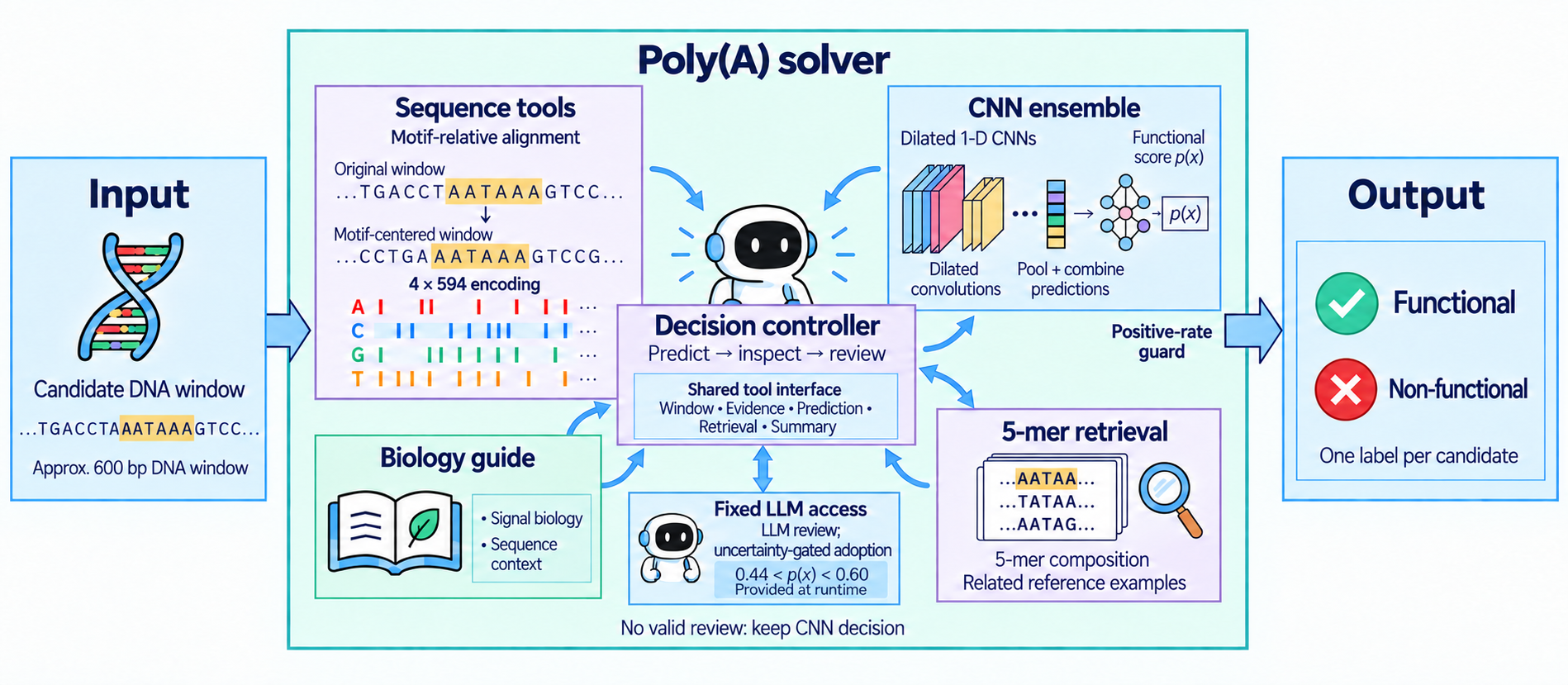}
\caption{\textbf{Internal structure of the polyadenylation solver.}
A candidate DNA window enters the solver and produces a functional or
non-functional prediction. The decision controller coordinates sequence tools, a
CNN ensemble, biological guidance, and 5-mer retrieval; curved arrows summarize
tool interaction and review at inference time. The LLM reviews every case; the
box labelled ``LLM review; uncertainty-gated adoption'' marks where its decision is adopted, only
inside the uncertainty band $0.44<p(x)<0.60$. The positive-rate guard caps
positives at 62\% of the batch before output (Eq.~\eqref{eq:v1-gate} and the four
stages in the text).}
\label{fig:v1-solver}
\end{figure}

The delivered solver processes a batch of cases in four stages
(Figure~\ref{fig:v1-solver}).
\begin{enumerate}
\item \emph{CNN score.} For every case the ensemble returns $p(x)$, the mean over
its four members and three input shifts (0 and $\pm4$ positions). The CNN
decision is $y_W(x)=\mathbb{1}[p(x)\ge 0.495]$, and a CNN-only prediction file is
written immediately so that the delivery is never empty.
\item \emph{LLM review.} For every case in the batch---no review deadline was
configured---the platform LLM, operating through the tool interface, inspects the aligned window, sequence evidence, the CNN score, and
retrieved neighbours, and returns a label $y_F(x)$.
\item \emph{Uncertainty-band arbitration.} The two decisions are combined as
\begin{equation}
\tilde y(x)=
\begin{cases}
 y_F(x), & \ell<p(x)<u\ \text{and a valid LLM answer is available},\\
 y_W(x), & \text{otherwise},
\end{cases}
\label{eq:v1-gate}
\end{equation}
with $(\ell,u)=(0.44,0.60)$. A missing, timed-out, or unparsable LLM answer
therefore leaves the CNN decision in place.
\item \emph{Positive-rate guard.} Let $n$ be the number of cases in the batch. If more than
$\lfloor 0.62\,n\rfloor$ cases are labelled positive after arbitration, the final
output labels exactly the $\lfloor 0.62\,n\rfloor$ cases with the highest CNN
score $p(x)$ as positive and all others as negative; otherwise $\tilde y$ is
output unchanged.
\end{enumerate}

\subsection{Complete-system result}
Independent evaluation of the complete Poly(A) solver recorded a positive-class F1 of
0.8977 on 3{,}000 cases (precision 0.8832, recall 0.9127), against a reference bar
of 0.88. The host passed all 3{,}000 cases to the solver in a single batch, so the
guard was applied once over the full set; the solver labelled 51.67\% of the cases
positive, below the 62\% cap, and the guard left every prediction unchanged. The delivered
artifact combines sequence processing, a fitted CNN ensemble, retrieval resources,
and a task-specific inference controller.
\FloatBarrier

\section{Discussion}
\label{sec:v1-analysis}
\subsection{Scientific system construction across tasks}
Across the cases, construction turns a task into an explicit representation and a
set of usable resources rather than a single prompt---a coordinate system for a
sequence, a trained predictor, a schema-search interface, a retrieval index---each
inspectable in the delivered code. This decomposition also clarifies attribution.
Because a solver pairs constructed resources with the platform LLM, the builder's
contribution cannot be read off the fraction of predictions the model changes at
inference: a strong numerical tool can be the main product of construction even
when the controller usually accepts its output, and conversely a task can be
carried mostly by the assembled environment. Separating the two---resources versus
controller---is what the builder/solver split makes possible.

Construction also builds its own feedback. The builder's development splits,
self-checks, and perturbed re-runs give a signal \emph{during} construction that is
distinct from the host's held-out evaluation; the worked example
(\S\ref{sec:v1-worked}) shows the resulting loop.

\subsection{Where construction contributes}
The reported cases show two forms of construction. In prediction tasks over sequences, structures, and signals, the builder develops the representation, prepares training data, and fits a task-specific predictor. The resulting predictive capability is retained in the delivered artifact and evaluated through its task metric.

In tool-driven tasks, construction assembles retrieval resources, executable tools, and answer checks around a fixed platform LLM. The recorded raw-LLM comparisons describe performance differences on the same inputs and metric; attributing those differences to construction would additionally require a controlled comparison with the same underlying model. The recorded baselines already clear the configured bars on all five compared subtests, so bar clearance alone does not establish an advantage from construction. Individual differences are reported in Appendix~\ref{sec:v1-matched-baselines}.

\subsection{Interpreting bar clearance}
Configured reference values serve as task-specific acceptance thresholds rather than uniformly matched reproductions of source results. Table~\ref{tab:v1-subtests} records their sources and any recomputation. Its notes identify differences in evaluation samples, data splits, input resolution, and reporting protocol, as well as unresolved training/test overlap. Clearing a configured bar establishes acceptance under the host protocol; it does not by itself establish an improvement over the source method under matched conditions.

\section{Conclusion}
\sys treats AI agents as builders of AI labs rather than only users of existing tools. What each construction leaves behind is a task-specific system with a fixed interface, together with the record that links how it was built to how it performs. Across the reported cases, the work is divided differently between constructed resources and inference-time control: in some labs a fitted predictor carries the result, in others the platform model operates tools and knowledge the builder assembled. Making that delivered lab, rather than the agent's development narrative, the object of study is what allows its components to be inspected and its outputs to be checked under a specified task protocol.

\FloatBarrier
\small
\bibliographystyle{plainnat}
\bibliography{refs}
\normalsize
\clearpage

\appendix

\section{Implementation and evaluation configuration}
\label{app:v1-protocol}
\paragraph{Builder and model endpoints.}
The reported runs use the archived builder identifier \texttt{claude-opus-5} with
the pi coding-agent harness, version 0.80.1. Host-mediated model endpoints in the
archive serve \texttt{claude-opus-5} and \texttt{kimi-k3}. The evaluation
records list the Opus endpoint as the solver's platform-LLM upstream for all 12
subtests. The raw-LLM baseline in Appendix~\ref{sec:v1-matched-baselines} directly queries
the host-configured baseline model on the evaluation inputs, without the constructed
components, and is recorded in the same result records. The baseline and solver
model endpoints are configured separately; whether they served the same model is not established by
the archived result records.

\paragraph{Runtime dependencies and limits.}
The build VM can access public data, packages, and model weights. Earlier batches
used a 12-hour wall-clock limit and later batches 24 hours; resource use was
metered without a common fixed budget across tasks.

\paragraph{Delivery interface.}
Each artifact exposes a single entry script: it reads a batch of cases---each with
an identifier and its task inputs---and writes a prediction file indexed by those
identifiers, carrying labels, scores, continuous estimates, SQL, or textual answers as
the task requires. The delivered environment may bundle trained weights, code,
retrieval resources, and knowledge files, and its inference path may call the
platform LLM through a host endpoint; it contains no newly trained LLM weights.
At evaluation the host remints identifiers, permutes input fields, and substitutes
a generic task description (\S\ref{sec:v1-boundary}).

\paragraph{Scoring methods.}
Primary metrics are computed by the host from the delivered predictions. Accuracy,
F1, AUROC, and Pearson $r$ are computed
against host-side references; SQL is scored by execution accuracy on the host
grading database, including the task's empty-answer convention. One model-graded metric type is used and kept separate from classification
accuracy. \emph{Judge accuracy} (bioinformatics analysis and clinical calculation):
a single host judge model receives the reference answer with its aliases and the
solver's answer, and returns whether the answer matches; judge accuracy is the
fraction of cases accepted. The judge model is taken from the host configuration,
with the baseline model as its default. The archived result records do not identify
which setting was used for these runs. The same records also report strict
string-match rates, a different metric: 0.2432 for bioinformatics and 0.8755 for
clinical calculation. This answer-level judge grades solver outputs. It is distinct from the
platform's process judge, which rates how an artifact was built and is not used in
this report.

\paragraph{Output checks.}
Every subtest requires predictions for at least half of the current identifiers
and at least $\min(2,n)$ distinct top-1 outputs; failure zeros that subtest.
Metric-specific rules can add constraints such as positive-rate guards. An
evaluation case without a prediction is scored as incorrect (0 for judge-graded
metrics). All 10 reported records passed these checks with predictions for every
evaluation case, i.e.\ full coverage on all 12 subtests.

\paragraph{Episode termination.}
Of the 10 reported runs, seven ended with an explicit submission. Three---polyadenylation,
protein--ligand affinity, and bioinformatics analysis---were closed by the operator
with a staged artifact after the builder stopped without submitting, and their staged
solvers were then evaluated through the same host-side path.

\section{Representative scientific constructions}
\label{app:v1-details}
This appendix records how each representative solver was constructed. Every case
is described by its input and output, the components the builder constructed
(stating which were fitted during construction), the data and development
resources it used, and its inference procedure. Development splits and
self-checks belong to construction; the reported values come only from independent evaluation (\S\ref{sec:v1-eval}). Where a solver runs an agent loop,
the recorded harness is pi; other paths implement a task-specific
observation--action loop over the same resources. The records describe what
enters the model context but do not measure per-case context length.

\subsection{Polyadenylation signal detection}
\paragraph{Input and output.}
A roughly 600-base DNA window containing a candidate polyadenylation hexamer; the
output is a functional or non-functional label.
\paragraph{Constructed components.}
Motif-relative canonicalization to a four-channel, 594-position input; an ensemble
of four dilated residual 1D CNNs, fitted during construction; a 5-mer composition
retrieval index; a biology knowledge card; and a shared tool interface exposing the
aligned sequence, sequence evidence, the CNN score, and retrieved examples.
\paragraph{Data and development resources.}
22,604 human windows from the public DeepGSR collection, split before training into
17,104 training, 2,500 validation, and 3,000 development-test windows; mouse and cow
sequences were available as optional augmentation. Validation guided ensemble
selection and the decision threshold. Self-checks ran the packaged solver with
fresh identifiers, shifted window boundaries, off-centre signals, and reordered
inputs.
\paragraph{Inference procedure.}
CNN scoring, LLM review, uncertainty-band arbitration, and a positive-rate guard,
as specified in \S\ref{sec:v1-worked}.

\subsection{Protein--ligand binding affinity}
\paragraph{Input and output.}
A three-dimensional protein--ligand complex; the output is a continuous affinity
estimate.
\paragraph{Constructed components.}
Geometry-based descriptors of the pocket and ligand (contact patterns, chemical
interaction descriptors, molecular shape); an ensemble of eight gradient-boosted
tree models, fitted during construction; a nearest-neighbour evidence index; and
tools exposing complex inspection, structural features, the learned estimate, and
ensemble disagreement.
\paragraph{Data and development resources.}
Public protein--ligand structures paired with affinity annotations, 5,311 training
examples in the delivered metadata. Development code supports a
sequence-similarity-based holdout. Structural self-checks rotate and translate
coordinates, reorder atoms, and rename chains or residues. The available metadata
do not verify that the CASF-2016 core set was excluded from the training data.
\paragraph{Inference procedure.}
The tree ensemble produces an affinity estimate; a platform-LLM tool loop examines
the structural evidence and proposes a refinement bounded around that estimate.
Without a usable review, the ensemble estimate is returned.

\subsection{Bioinformatics analysis}
\paragraph{Input and output.}
A study data package and a scientific question; the output is an answer computed
from the supplied data.
\paragraph{Constructed components.}
An executable analysis environment with the required scientific software;
reusable procedures for table inspection, differential expression, enrichment
analysis, gene-identifier mapping, regression, principal-component analysis, and
phylogenetic distances; domain recipes covering contrast direction,
multiple-testing correction, enrichment backgrounds, and output units; offline
gene-set resources; and a data summarizer reporting file shapes, column types,
representative values, and numeric ranges. No predictive model is fitted during
construction.
\paragraph{Data and development resources.}
Development questions and reference answers reconstructed from public BixBench
study packages. The complete workflow was run with changed identifiers and
reordered inputs, with numeric, textual, and model-based answer comparisons.
\paragraph{Inference procedure.}
A pi coding agent works in a case-specific workspace: it reads the data summary,
selects and executes an analysis, and formats the answer. Repeated sample files
are summarized by representative profiles, and the harness compacts context in
longer sessions.

\subsection{Clinical text-to-SQL}
\paragraph{Input and output.}
A clinical database schema and a natural-language question; the output is an SQL
query or an abstention for an unsupported request.
\paragraph{Constructed components.}
Schema search, column-encoding inspection, clinical synonym support,
answerability checks, query-execution checks, and two schema-faithful synthetic
databases. No predictive model is fitted during construction.
\paragraph{Data and development resources.}
Two populated development databases with eICU-like and MIMIC-like structures and
about 350 template-generated questions with derived reference queries, covering
counts, joins, aggregation, time windows, outcomes, and unanswerable requests.
Self-checks shuffle schema order and compare returned row sets, separating invalid
queries from correct abstentions.
\paragraph{Inference procedure.}
A pi tool loop, with a task-specific controller as an alternative path, inspects
tables, drafts candidate queries, executes them on the synthetic databases, and
revises them from the results; candidate trajectories are compared by their
agreement on synthetic instances before a query is returned.

\subsection{Clinical calculation}
\paragraph{Input and output.}
A patient note and a calculation question; the output is a score, numerical value,
or date.
\paragraph{Constructed components.}
A catalogue of 55 clinical calculators with specifications, parameter schemas, and
executable rules; unit conversion, rule tables, and rounding; evidence search over
the note; per-criterion audit tools; and a deterministic extraction fallback. No
predictive model is fitted during construction.
\paragraph{Data and development resources.}
Calculator checks pass reference parameters directly to each calculator;
end-to-end checks vary question phrasing, field order, identifiers, and unit
spellings. Error analysis refined evidence attribution, negation handling, and
parameter normalization.
\paragraph{Inference procedure.}
The platform LLM identifies the calculation from its specification card, extracts
the parameters with supporting evidence, and calls the calculator; tool feedback
returns the parameters used and a derivation, and unit and plausibility checks flag
inconsistent inputs for re-extraction. Arithmetic is done by the executable rules.

\subsection{Chemistry reasoning}
\paragraph{Input and output.}
A chemistry question with answer options; the output is an answer choice.
\paragraph{Constructed components.}
Molecular and formula-processing tools, chemical-arithmetic and numerical
utilities, and topic-specific reference material. No predictive model is fitted
during construction.
\paragraph{Data and development resources.}
Development questions from public chemistry question sets, with separate slices
for refining instructions and for end-to-end checks. Checks vary option order,
question formatting, and identifiers, and separate incorrect choices from
answer-parsing failures.
\paragraph{Inference procedure.}
Computed properties and composition summaries are added to the model's
observations where the question contains an interpretable structure or formula. A
tool-driven attempt is checked against additional reasoning passes; disagreement
triggers further verification, and the controller aggregates the choices.
Interaction depth adapts to the remaining execution budget, with a shorter path
when a longer rollout cannot finish.

\subsection{Protein interaction}
\paragraph{Input and output.}
Two amino-acid sequences; the output is a real-valued interaction score for
ranking pairs.
\paragraph{Constructed components.}
Frozen ESM-2 embeddings of each sequence combined with physicochemical descriptors
in a symmetric pair predictor, fitted and calibrated during construction; tools
exposing charge, hydropathy, composition, and disorder propensity; a
training-derived interaction knowledge resource; and a domain playbook.
\paragraph{Data and development resources.}
About 1.46 million labelled pairs over about 83,000 proteins from public interaction
collections, normalized, deduplicated as unordered pairs, cleared of contradictory
labels, and balanced by source and class. Proteins were partitioned before forming
training, validation, and development-test pairs, and candidate predictors were
selected on per-source validation performance. Self-checks exchange pair order,
alter input formatting, and simulate agent-call failures.
\paragraph{Inference procedure.}
The predictor scores the pair; a pi review of the biophysical evidence applies a
bounded adjustment. The learned score is returned when review cannot complete.

\section{Complete task results and baseline comparisons}
\label{app:expanded-subtests}
Table~\ref{tab:v1-subtests} lists every subtest of the 10 reported cases, which
comprise 12 subtests, with its metric type, number of evaluated cases, recorded
value, and configured reference. Margins over the reference vary widely; the
smallest are polyadenylation signal detection (0.8977 vs.\ 0.88) and 3D abdominal
organ classification (0.9705 vs.\ 0.952).
\begingroup
\scriptsize
\setlength{\tabcolsep}{3pt}
\begin{longtable}{@{}>{\raggedright\arraybackslash}p{3.3cm}>{\raggedright\arraybackslash}p{2.0cm}>{\raggedright\arraybackslash}p{1.35cm}rrr>{\raggedright\arraybackslash}p{2.9cm}@{}}
\caption{All subtests of the 10 reported cases, comprising 12 subtests. Solver is the delivered solver's value under independent evaluation and Reference bar the configured reference value, both on the primary metric. These metric types (Appendix~\ref{app:v1-protocol}) are reported separately and should not be interpreted interchangeably; judge accuracy and execution accuracy are not classification accuracies. Reference-source markers and \mbox{notes 1--6} are explained below the table.}\label{tab:v1-subtests}\\
\toprule Task & Subtest & Metric & $n$ & Solver & \shortstack[r]{Reference\\bar} & Reference source\\ \midrule
\endfirsthead
\toprule Task & Subtest & Metric & $n$ & Solver & \shortstack[r]{Reference\\bar} & Reference source\\ \midrule
\endhead
\bottomrule\endfoot
Protein--protein interaction & Bernett PPI & AUROC & 3000 & 0.9874 & 0.7000 & \citealp{liu2025plminteract}$^{4}$\\
Protein--ligand binding affinity & PDBbind & Pearson $r$ & 285 & 0.9141 & 0.8030 & \citealp{graber2025resolving}$^{3}$\\
Bioinformatics data analysis & BixBench & Judge acc. & 296 & 0.4561 & 0.2280 & \citealp{ghareeb2026multiagent}$^{1}$\\
Polyadenylation signal detection & Poly(A) signal & F1 & 3000 & 0.8977 & 0.8800 & \citealp{shen2026foundation}$^{5}$\\
Abdominal organ classification (3D CT) & Organ 3D & Accuracy & 610 & 0.9705 & 0.9520 & \citealp{liu2026manifold}$^{6}$\\
Chemistry reasoning & ChemBench & Accuracy & 942 & 0.9268 & 0.8482 & \citealp{mirza2025framework}$^{\ddagger}$\\
Medical reasoning & Medbullets & Accuracy & 308 & 0.9123 & 0.6331 & \citealp{chen2025benchmarking}\\
Medical reasoning & MedMCQA & Accuracy & 4183 & 0.8735 & 0.7366 & \citealp{nori2023capabilities}\\
Medical reasoning & MedQA & Accuracy & 1273 & 0.9678 & 0.9020 & \citealp{nori2023can}\\
Clinical calculation & MedCalc & Judge acc. & 1100 & 0.9445 & 0.8500 & \citealp{krohngrimberghe2026medcalc}$^{2}$\\
Clinical text-to-SQL & EHRSQL & Exec.\ acc. & 1000 & 0.5470 & 0.3200 & \citealp{bedi2025medhelm}$^{\dagger}$\\
Consumer-health answer verification & MEDIQA & Accuracy & 1107 & 0.8193 & 0.7800 & \citealp{benabacha2019overview}\\
\end{longtable}
\endgroup

\paragraph{Notes to Table~\ref{tab:v1-subtests}.}
$^{\dagger}$~Public leaderboard entry (MedHELM v4.0.0, GPT-4o) rather than a value
reported in a paper.
$^{\ddagger}$~Recomputed: ChemBench 0.8482 is o1-preview accuracy recomputed from the
released per-question scores on the 942-question subset evaluated here and does not
appear in the paper.
\textsuperscript{1}~BixBench: 0.228 is the Finch agent's accuracy on a 170-question
expert-selected panel reported by \citet{ghareeb2026multiagent}, not on the 296
BixBench questions \citep{mitchener2025bixbench} evaluated here.
\textsuperscript{2}~MedCalc: the configured value 0.85 is not reported as such; the nearest
reported result is 85.5\% (open-book GLM-4.7) on a 275-case subsample, and the best
open-book result on all 1{,}100 cases is 81.5\% \citep{krohngrimberghe2026medcalc};
benchmark from \citet{khandekar2024medcalc}.
\textsuperscript{3}~Protein--ligand affinity: training/core-set overlap not verified.
\textsuperscript{4}~Protein interaction: the host evaluates a seeded, stratified
subsample of 3{,}000 pairs from the approximately 52{,}700-pair Bernett test split.
The reference 0.70 is the published PLM-interact result on the source test split,
not a recomputation on these 3{,}000 pairs.
\textsuperscript{5}~Poly(A): host evaluation uses 3{,}000 class-balanced windows
sampled with seed 130 from the DeepGSR-derived, 600-base sequences distributed by
DeepGenGrep. This sampling does not reproduce the source paper's evaluation split
for the reference 0.88. The builder's development split described in
\S\ref{sec:v1-worked} is separate; sample disjointness between its acquired data
and host evaluation has not been established.
\textsuperscript{6}~Organ 3D: host inputs are $28\times28\times28$ volumes and the
solver result is from one build. The reference 0.952 from \citet{liu2026manifold}
uses $64\times64\times64$ inputs and averages three random seeds.

\FloatBarrier
\subsection{Recorded accuracy baselines}
\label{sec:v1-matched-baselines}
\begin{figure}[!ht]
\centering
\includegraphics[width=\linewidth]{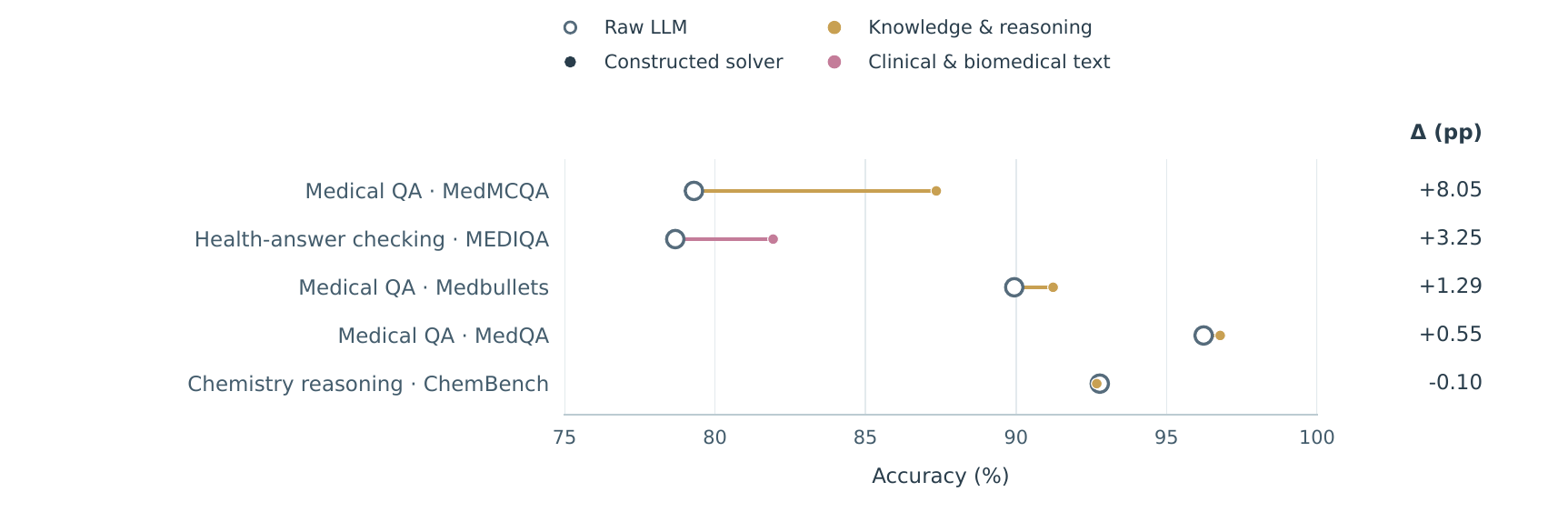}
\caption{\textbf{Constructed solvers versus the raw-LLM baseline.}
All five accuracy comparisons on the same inputs from three cases, sorted by the
solver-minus-baseline difference. Open circles denote the raw LLM and filled
circles the constructed solver; colors follow the scientific categories in
Figures~\ref{fig:v1-dumbbell} and~\ref{fig:v1-resources}. The right column reports changes in percentage points
(pp); model equivalence is unverified. Row labels give an abbreviated task name and the subtest; ``Medical QA''
is the medical reasoning task of Table~\ref{tab:v1-subtests}.}
\label{fig:v1-baseline}
\end{figure}

Figure~\ref{fig:v1-baseline} and Table~\ref{tab:v1-baselines} report every
accuracy-scored subtest for which a recorded raw-LLM baseline is available: five
subtests from three of the reported cases, each evaluated on the same held-out inputs
and metric for both systems. These comparisons match the evaluation inputs and metric,
but do not establish model equivalence (Appendix~\ref{app:v1-protocol}). The solver
scores higher on four subtests and lower on ChemBench ($-0.10$~pp). On all five
subtests the recorded raw-LLM baseline already exceeds the configured reference.

\begin{table}[!ht]
\centering
\caption{The five recorded accuracy comparisons plotted in
Figure~\ref{fig:v1-baseline}, in the same order. Raw LLM and Solver are accuracies
on a 0--1 scale; $\Delta$ is Solver minus Raw LLM in percentage points.}
\label{tab:v1-baselines}
\begingroup\small
\begin{tabular}{@{}lrrrr@{}}
\toprule Subtest & $n$ & Raw LLM & Solver & $\Delta$ (pp)\\ \midrule
MedMCQA & 4183 & 0.7930 & 0.8735 & +8.05\\
MEDIQA & 1107 & 0.7868 & 0.8193 & +3.25\\
Medbullets & 308 & 0.8994 & 0.9123 & +1.29\\
MedQA & 1273 & 0.9623 & 0.9678 & +0.55\\
ChemBench & 942 & 0.9278 & 0.9268 & -0.10\\
\bottomrule
\end{tabular}
\endgroup

\end{table}
\FloatBarrier

\end{document}